%% file: Afser2020.tex
\documentclass[journal]{IEEEtran}
\usepackage{hyperref}
\usepackage[utf8]{inputenc}
\usepackage{graphicx}
\usepackage{amsmath}
\usepackage{amsthm}
\usepackage{amsmath, amsthm, amssymb}
\usepackage{epstopdf}
\usepackage[mathscr]{eucal}  
\usepackage{float} 
\usepackage{multicol}
\usepackage{caption}
\usepackage{subcaption}
\usepackage{enumerate}
\usepackage{authblk}
\usepackage{multirow}
\usepackage{tikz}
\usetikzlibrary{decorations.pathreplacing}
\usepackage{multirow}
\usepackage{cite}
\usepackage[ruled,vlined]{algorithm2e}
\SetAlFnt{\footnotesize}
\SetKw{KwBy}{by}
\usepackage[noend]{algpseudocode}

\makeatletter
\def\BState{\State\hskip-\ALG@thistlm}
\makeatother

\usetikzlibrary{positioning}

\usepackage{tikz}
    \usetikzlibrary{shapes.geometric}   
\usepackage{rotating}

\usepackage{mathtools}
\usepackage{url}
\usepackage{amssymb}
\usepackage{physics}

\newtheorem{corollary}{Corollary}

\renewcommand{\theenumi}{\roman{enumi}}
\renewcommand{\vec}{\vb*}

\begin{document}
%


\title{A Baseline Statistical Method For Robust User-Assisted Multiple Segmentation}
\author{H\" useyin~Af\c ser \thanks{ \hspace{- 11.2pt} The submission date is 03/01/2022. \newline \hspace{18pt} H\" useyin~Af\c ser is with the Adana Alparslan T\"{u}rke\c{s} Science and Technology University, Department  of Electrical Electronics Engineering, 01250, Adana, Turkey. (e-mail:afser@atu.edu.tr)}}

\maketitle

\begin{abstract}

Recently, several image segmentation methods that welcome and leverage different types of user assistance have been developed.
In these methods, the user inputs can be provided by drawing bounding boxes over image objects, drawing scribbles or planting seeds that help to differentiate between image boundaries or by interactively refining the missegmented image regions. Due to the variety in the types and the amounts of these inputs, relative assessment of different segmentation methods becomes difficult. As a possible solution, we propose a simple yet effective, statistical
segmentation method that can handle and utilize different input types and amounts. The proposed method is based on robust hypothesis testing, specifically the DGL test, and can be implemented with time complexity that is linear in the number of pixels and quadratic in the number of image regions. Therefore, it is suitable to be used as a baseline method for quick benchmarking and assessing the relative performance improvements of different types of user-assisted segmentation algorithms. We provide a mathematical analysis on the operation of the proposed method, discuss its capabilities and limitations, provide design guidelines and present simulations that validate its operation.

\end{abstract}


\vspace{-10 pt}
\section{Introduction}

In image segmentation the aim is to assign a label to every image pixels in a way that the pixels with the same labels exhibit similar or unique properties. These properties can be low-level features such as color and texture or high-level semantic descriptions. 

Image segmentation, in general, is a challenging problem. Therefore, depending upon the target application
and specifications there exist several approaches to segmentation that differ in capabilities and limitations \cite{Gonzalez}. To ease this non-trivial task, recently numerous methods that accept and utilize different types of user assistance have been developed. For example in Graph Cuts \cite{GraphCut}, Random Walk \cite{RandamWalk} and their many variants \cite{51,54,57,59,65,62,66,161,162}, the user inputs are seed points or scribbles. In GrabCut \cite{Grabcut} and its variants \cite{81,83, 80, 85} the user inputs are bounding boxes where these methods may also incorporate additional user inputs to refine the segmentation process. A detailed survey on user-assisted segmentation methods can be found in \cite{Survey}.

The variant algorithms cited above are often developed with the intention of improving the performance of the original method \cite{51,54,57,59,62,65,80,161}, decreasing or easing the human interaction \cite{80,66,59,83,65} or providing robustness to the user input \cite{54, 81, 83, 162}. However, a great majority of these methods are limited to binary, i.e. foreground and background, segmentation. Besides, due to their construction and inherent operating mechanism, they work best under a specific input type and their performances improve with the amount or the precision of the user inputs. Therefore, it becomes quite difficult to assess the relative performance improvements of these methods within a unified framework. This difficulty only increases if their varying complexities are taken into the assesment picture.

With the intention of providing a common ground to different user assisted segmentation methods, we present a novel, minimalistic, multiple segmentation algorithm based on robust hypothesis testing \cite{HuberA, Gul} and specifically the DGL test \cite{Devroye, Biglieri,Afser}. Being algorithmically simple, robust to different input types and requiring a complexity that is linear in the number of pixels, the proposed method is suitable to be used as a baseline method for quick benchmarking purposes. The main contributions of this paper are:

\textit{1)} We show that one can take advantage of the inherent robustness of the  DGL test in user-assisted image segmentation. Some user input types that we consider are:

\textit{i)} a fraction of labeled pixels within ground truths masks or bounding boxes

\textit{ii)} randomly selected seed points within ground truths masks

\textit{iii)} randomly perturbed bounding boxes

\textit{2)} The proposed method is algorithmically simple and can be coded with around 30 Matlab sentences. It can be efficiently implemented with a time complexity that is linear in the number of pixels and quadratic in the number of image labels.

\textit{3)} We provide a mathematical analysis on the operation of the proposed method. 

\textit{4)} We present simulations on Berkeley's BSDS 500 database \cite{Berkeley} that validate the robust operation of the proposed method.

\textit{5)} In addition to the initial user input, the increases in the accuracy of the proposed method per additional user inputs can easily be benchmarked where the accuracy can be
increased arbitrarily.

\vspace{-15 pt}
\section{Preliminaries}\label{sect:sect_2}
\vspace{-8 pt}
\subsection{Problem Statement}

Consider segmenting an image $I \hspace{-4 pt} :\Omega \rightarrow \mathbb{R}^d$ with $N$ pixels,  $|\Omega| = N$, into $M$ disjoint and collectively exhaustive regions $\Omega_1, \Omega_2,...,\Omega_M$ where a region may consist of several topologically separated components. Let $X$ denote an arbitrary pixel in $\Omega$ and $I(X)$ be its intensity. We adopt the probabilistic formulation in \cite{Kim} and assume that pixel intensities are independent and identically distributed (i.i.d.) and these distributions are different in $\Omega_i$, $i=1,2,...,M$. More formally
\begin{align}
\{ I(X) | X \in \Omega_i \} \sim P_i, \quad \quad i=1,2,...,M,
\end{align}
where, in general,  $P_i:\mathbb{R}^d \rightarrow [0,1]$. The assumed model is depicted in Figure 1. For grayscale images $d=1$ and for RGB, HSV, etc., images $d=3$. We will also consider digital images where pixel intensities are quantized along $d$ dimensions to form a discrete product alphabet ${\cal X} = {\cal X}_1 \times {\cal X}_2 \times....\times{\cal X}_d $ so that $I : \Omega \rightarrow {\cal X} $ and $P_i : {\cal X} \rightarrow [0,1]$.

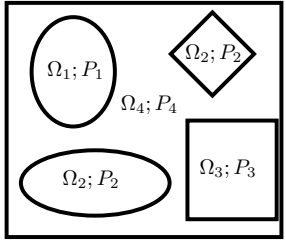
\begin{figure}
\begin{center}
\input{fig_0.tex}
\end{center}
\caption{Illustration of the assumed image model for the case $M=4$. The image consists of 4 regions $\Omega_1,\Omega_2, \Omega_3,\Omega_4$ where the distributions of pixel intensities in each region are distinct.}
\label{fig:fig_1}
\end{figure}

The aim in segmentation is to devise a labeling rule  $L : \Omega \rightarrow \{ 1, 2,...,M \}$ such that $L(X)=i$ is chosen if $X \in \Omega_i$.

\vspace{-10 pt}
\subsection{DGL Test}

The DGL test \cite{Devroye} is a robust $M$-ary hypothesis testing procedure for i.i.d. sequences. It 
can be used when the true distributions, $P_i$, $i=1,2,...,M$, that are responsible for the generation of an observed sequence is not known, but there exist nominal distributions $Q_i$, $Q_i : \mathbb{R}^d \rightarrow [0,1]$, that are close to true ones. The test is robust provided that there exists 
a positive $\Delta$ such that  
\begin{align}
V(P_i,Q_i) \leq (\min_{ i \neq j} V(Q_i,Q_j) -\Delta)/2 \label{eq:DGL_robustnes}
\end{align}
where $V(P,Q)$ denotes the total variation between $P$ and $Q$. If the condition in \eqref{eq:DGL_robustnes} is satisfied, the resultant test will have a probability of error that decreases exponentially in the length of the test sequence uniform for all $P_i$. 

Recently, we have investigated the DGL test in the discrete setting for statistical classification with labeled training sequences (see \cite{Afser}). Our findings indicate that the achievable error exponent of the DGL test is sub-optimal, however it provides consistent classification for any length of training sequences provided that the unknown sources are separated in total variation. We will use the following corollary \cite[Corollary 1,2]{Afser} to analyze the performance of the proposed  method.
\begin{corollary}
Let $\vec{t}^{n_t}_i=[ t_{i1}, t_{i2},...,t_{in_t}]$, $i=1,2,...,M$, be a set of training sequences such that $\vec{t}^{n_t}_i$ is generated from distribution $P_i$. In the discrete setting where  $P_i : {\cal X} \rightarrow [0,1]$ and 
when the empirical distribution  of $\vec{t}^N_i$ is taken as the nominal distribution $Q_i$ in the DGL test, the error probability for testing a sequence with length $n$  is upper bounded as
\begin{align}
\hspace{-12 pt} \Pr [e]   \leq 2Me^{-n ( \frac{ \alpha \min_{i \neq j}V(P_i,P_j)^2 }{2(2 + \sqrt{\alpha})^2} - \max \{ \frac{2 \ln (M-1)}{n}, \frac{| {\cal X} | \ln 2}{n} \} )} \label{eq:thm_1}
\end{align}
and
\begin{align}
\hspace{-11 pt} \Pr [e]   \leq  2M e^{-n ( \frac{ 2\alpha \min_{i \neq j}V(P_i,P_j)^2}{{(3| \cal X|}+2\sqrt{\alpha})^2}-  \max \{ \frac{2 \ln (M-1)}{n}, \frac{\ln |{\cal X}|}{n} \} ) }\label{eq:thm_2}
\end{align}
\vspace{-5 pt}
where $\alpha = n_t/n$.
\end{corollary}

The above bounds are not tight in general, however their non-asymptotic nature helps in characterizing the error exponent for finite $n$ and $\alpha$.

\vspace{-10 pt}
\section{ DGL Test-Based Segmentation}\label{sect:sect_3} 

\vspace{-5 pt}
\subsection{Implementation}\label{sect:sect_3A}

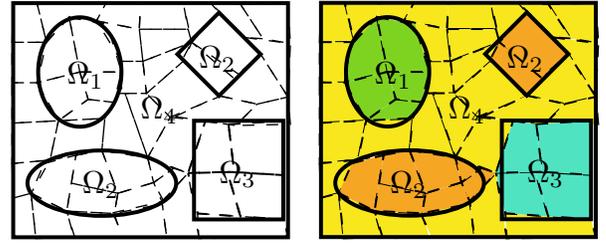
\begin{figure}[t!]%
    \centering
    \subfloat[]{ \input{fig_1.tex} }%
    \subfloat[]{\input{fig_2.tex}}%
    \caption{a) Dashed lines represent superpixel boundaries. b) Performing image segmentation at the superpixel level.}%
    \label{fig:example}%
\end{figure}

Let ${\cal  T}_1, {\cal  T}_2,....,{\cal  T}_M$ be $M$ sets of user-labeled image pixels. These pixels may be obtained from the ground truth masks,  the bounding boxes or from the vicinity of image seeds/scribbles as we explain in detail in the next section. Let $\hat{Q}_i : {\cal X} \rightarrow [0,1]$, $ {\cal X} ={\cal X}_1 \times {\cal X}_2 \times ... \times {\cal X}_d$, $i=1,2,....,M$, denote the possibly quantized intensity histograms of ${ \cal  T}_i$.
More formally,
\begin{align}
\hat{Q}_i (q)_{q \in {\cal X} } = \frac{1}{ | {\cal T}_i |}\sum_{ X \in {\cal T}_i} \mathbb{I}_{  \bar{I}(X) = q}
\end{align}
where $\mathbb{I}$ is an indicator function and $\bar{I}(X)$ denotes the rounded intensity of pixel $X$ where rounding is towards the nearest edge of the histogram bin descriptor.

We perform image labeling on the superpixel level by first partitioning $\Omega$ into superpixels, $\Omega \rightarrow \{ {\cal S}_1,{\cal S}_2,....,{\cal S}_K \}$, $ K \leq N$, and then apply the DGL test for each superpixel. The proposed approach is illustrated in Figure 2. Although the premise of superpixels is their boundary adherences, in practice this may not always hold and assigning the image labels at the superpixel level may result in some residual error as can be seen in Figure 2b. However, their adoption simplifies the proposed approach as they group adjacent pixels into perceptually similar cells that are easy to differentiate with the DGL test.  

In order to apply the DGL test, one needs to calculate $M(M-1)/2$ Borel set $A_{i,j}$, $ A_{i,j} \in \cal A$, that are of the form
\begin{align}
A_{i,j} = \left \{ q : \hat{Q}_i(q) \geq \hat{Q}_j(q) \right \},  \quad 1 \leq i < j \leq M, \label{eq:Borel_sets}
\end{align}
for all $q \in {\cal X}$. Then, if we let $ \{ X_1,X_2,...,X_n \}$ be the set of pixels in ${\cal S}_k$, the DGL test decides
$\hat{L}({\cal S}_k) =i$ if
\begin{align}
\hspace{-8 pt}\max_{A \in \cal {A} }  \left | \int_{A } \hspace{-5 pt} \hat{Q}_i(A) -\hspace{-2 pt} \mu_n( A) \right|= \hspace{-10 pt} \min_{j=1,...,M}  \max_{A \in \cal {A} }  \left | \int_{A } \hspace{-5 pt} \hat{Q}_j(A) -\mu_n( A) \right|\label{eq:DGL_test_rule}
\end{align}
where $ \mu_n( A) = \frac{1}{n} \sum_{i=1}^{n} \mathbb{I}_{X_i \in  A}$ and integrals in \eqref{eq:DGL_test_rule} can be calculated numerically over ${\cal X}$.

The pseudo-code for implementing the proposed segmentation algorithm in Matlab for the case $d=3$ is presented in Algorithm 1. This algorithm is minimalistic in the sense that 
it is based on 4 simple functions and can be coded with around 30 sentences. 

\vspace{-10 pt}
\subsection{Complexity}\label{sect:sect_3} 

The superpixels ${\cal S}_1, {\cal S}_2,...., {\cal S}_K$ can be calculated with complexity $O(N)$ e.g. by using SLIC algorithm \cite{SLIC} in Matlab. The calculations of $\hat{Q}_1,\hat{Q}_2,....,\hat{Q}_M$ can be done in $O(N)$, as well. Investigating Algorithm 1, we observe that the main loops in \textit{DGLSegment} can be calculated with time complexity $O(M^2 N + M^2 \log M)$ and require a space complexity of $O(M^3 N)$.  The initialization functions $Calc_{\cal A}$ and $Prob_{\cal A}$ can be calculated with time complexity $O(M^2 |{\cal X}|)$, ${\cal X} = {\cal X}_1 \times {\cal X}_2 \times ... \times {\cal X}_d$, and require space complexities $O(M^2 |{\cal X}|)$ 
and  $O(M^3 |{\cal X}|)$, respectively. Therefore, when $M$ is not exponential in $N$, the proposed method can be implemented with
time complexity $O(M^2\max \{ |{\cal X}|, N \})$ and space complexity $O(M^3\max \{ |{\cal X}|, N \})$. When $ |{\cal X}| >N$  the complexities are not linear in the number of pixels
which may be undesired. In such cases, one can apply dimensionality reduction and implement the algorithm on  $ \cal X'$,
 ${\cal X'} = {\cal X}_1' \times {\cal X}_2' \times ... \times {\cal X}_{d'}'$, $d' \leq d$, by forcing $|{\cal X'}| \leq  N$.  This ensures an algorithm with time complexity $O(M^2 N)$ and space complexity $O(M^3 N)$. Imposing dimensionality  reduction may result in some performance degradation; however, as we show via simulations in the next section this degradation may be negligible.

\begin{algorithm}[t!]

 \SetKwFunction{FMain}{$DGLSegment$}
 \SetKwProg{Fn}{Function}{}{}

\Fn{\FMain}{
\textbf{input}
 ${\cal S}_1,{\cal S}_2,...,{\cal S}_K$, $\hat{Q}_1,\hat{Q}_2,....,\hat{Q}_M$

\textbf{output: } $\hat{L}({\cal S}_1),\hat{L}({\cal S}_2),....,\hat{L}({\cal S}_K)$
$ {\cal A}=Calc_{\cal A}(\hat{Q}_1,....,\hat{Q}_M)$,  
$P_A=Prob_{\cal A}(\hat{Q}_1,...,\hat{Q}_M)$
 
 \For{$k=1:K$}
 {
$X={\cal S}_k$; 
  
  \For{$i=1:M-1$}
  {
    \For{$j=i+1:M$}
  {
  	$ \mu_A = Mean_n(X,A(:,:,:,i,j))$;
	
	    \For{$m=1:M$}
 		 {
		$d(m,i,j) = abs ( P_A(m,i,j)-\mu_A)$;
		$t(m) = max( max(d(m,:,:));$
	  }
    }
    }
    $ \hat{L}(S_k) = argmin(t)$;
  }  
    \KwRet\
    }
\SetKwFunction{FMain}{$Calc_{\cal A}$}
 \SetKwProg{Fn}{Function}{}{}
\Fn{\FMain}{
\textbf{input: } $\hat{Q}_1,\hat{Q}_2,....,\hat{Q}_M$ \textbf{output: }${\cal A}$

  \For{$i=1:M$}
  {
  	  \For{$j=i:M-1$}
	  {
	    [$in_1, in_2, in_3$] =find( $\hat{Q}_i(:,:,:) > \hat{Q}_j(:,:,:)$ )
	    
	    \For{$t=1:length(in_1)$}
	  {
	 	 $A(in_1(t),in_2(t),in_2(t),i,j)=1$;
	  }
	}
    }
        \KwRet\
}

\SetKwFunction{FMain}{$Prob_{\cal A}$}
 \SetKwProg{Fn}{Function}{}{}
\Fn{\FMain}{
\textbf{input: } $\hat{Q}_1,\hat{Q}_2,....,\hat{Q}_M, {\cal A}$

\textbf{output: }  $ \sum_{A } \hat{Q}_i(A), \forall A \in {\cal A}$, $i=1,2,...,M$ 

  \For{$m=1:M$}
  {
  	  \For{$i=i:M-1$}
	  {

	    \For{$j=i:M$}
	  {
	 	 \hspace{-2 pt}$P_A(m,i,j)=trapz(A (:,:,:,i,j).*Q_m(:,:,:))$\tcp*[l]{Trapezoidal numerical integration over ${\cal X}_1 \times {\cal X}_2 \times {\cal X}_3$}
	  }
	}
    }
        \KwRet\
}

\SetKwFunction{FMain}{$Mean_n$}
 \SetKwProg{Fn}{Function}{}{}
\Fn{\FMain}{

\textbf{input:} $ A \in \cal A$, $X$ \textbf{output:}  $\mu_n(A)$,   $n=length(X)$;  

 \For{$i=1:n$}
  {
  	  \If{$A(X(i,1),X(i,2),X(i,3))==1$}{
	  $I=I+1;$
	  }
    }
$ \mu_n=I/n;$   
    \caption{ Robust User-Assisted Segmentation}

}

\end{algorithm}

\vspace{-10 pt}
\subsection{Performance}\label{sect:sect_3A}

In addition to the adopted probabilistic model in Section 2A, let us also assume that the superpixels adhere perfectly to the boundaries as ${\cal S}_k \subset \Omega_i$, for some $i=1,2,...,M$, $k=1,2,...,K$.
Then, the following result is a consequence of Corollary 1.
\begin{corollary}
Let $n_{min} = min\{ | {\cal T}_1| , |{\cal T}_2|,...., |{\cal T}_M| \}$ by assuming the pixels  are selected from ground truth masks.
Also, let ${\cal X} = {\cal X}_1 \times {\cal X}_2...\times {\cal X}_d$ be the product alphabet for the image intensities i.e. $I : \Omega \rightarrow {\cal X}$. Then, the probability of mislabeling ${\cal S}_k = \{ X_1,X_2,....,X_n \}$ with the proposed method obeys the upper bounds in \eqref{eq:thm_1} and \eqref{eq:thm_2} by letting $\alpha= n_{min}/n$.
 \end{corollary}

The bounds in $\eqref{eq:thm_1}$ and $\eqref{eq:thm_2}$ imply that the error probability decreases exponentially in the size, $n$, of the superpixel. Therefore, the image should be decomposed into superpixels whose sizes are as large as possible. However, due to the imperfect boundary adherences of the superpixels and the desired granularity in the image, the number of superpixels should be adjusted so that there exist a few of them in the smallest $\Omega_i$. Next, we observe that the exponent term in  \eqref{eq:thm_1} is positive provided that
\begin{align}
n  \geq  \frac{  2 (1+\sqrt{\alpha} )^2 |{\cal X} | \ln (2) }{ \alpha \min_{i \neq j} V( P_i, P_j )^2}
\end{align}
which implies that the performance depends on the relative sizes of the alphabet and the superpixels, as well. If $|\cal X|$ scales faster than $n$, the bound in \eqref{eq:thm_2} can be used and this 
ensures a positive error exponent provided that $\sqrt{\alpha}$ scales faster than $|{\cal X}|$ as $n_{min} \geq |{\cal X}|^{2} n$. Therefore, by increasing the amount of user provided labeled pixels from each $\Omega_i$ one can ensure the correct operation of the proposed test for any $|{\cal X}|$ and $n$.

When the user inputs are labeled pixels from bounding boxes, $\hat{Q}_i$ is no longer an empirical density of $P_i$ but a mixture. For this case, the proposed method is still robust
with a positive error exponent provided that $ V(P_i,\hat{Q}_i) \leq ( \min_{ i \neq j} V(\hat{Q}_i,\hat{Q}_j) -\Delta)/2 $ for some positive $\Delta$. However, as the bounding boxes get loose, the mixing of distributions increases and thus  $\min_{ i \neq j} V(\hat{Q}_i,\hat{Q}_j)$ decreases. This, in turn, degrades the performance of the proposed method.

\begin{figure*}

%
%

\begin{center}
\begin{minipage}[t]{0.135 \linewidth}
 \includegraphics[scale=0.145]{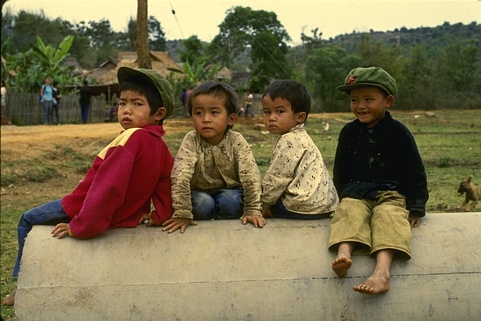}
\end{minipage}
\begin{minipage}[t]{0.135 \linewidth}
\includegraphics[scale=0.145]{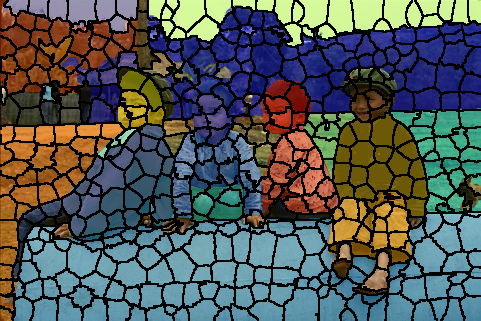}
\end{minipage}
\begin{minipage}[t]{0.135 \linewidth}
\includegraphics[scale=0.145]{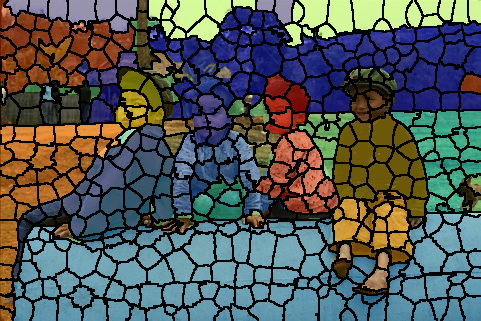}
\end{minipage}
\begin{minipage}[t]{0.135 \linewidth}
\includegraphics[scale=0.145]{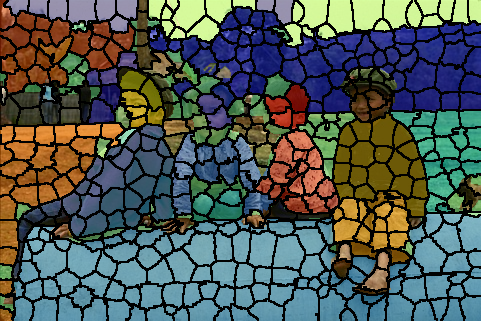}
\end{minipage}
\begin{minipage}[t]{0.135 \linewidth}
\includegraphics[scale=0.145]{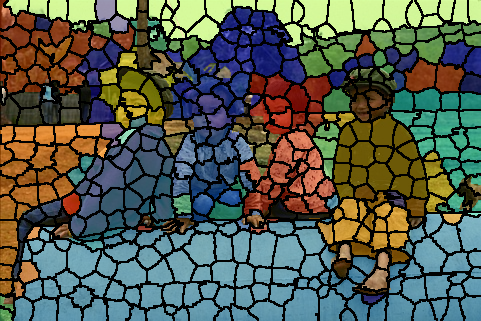}
\end{minipage}
\begin{minipage}[t]{0.135 \linewidth}
\includegraphics[scale=0.145]{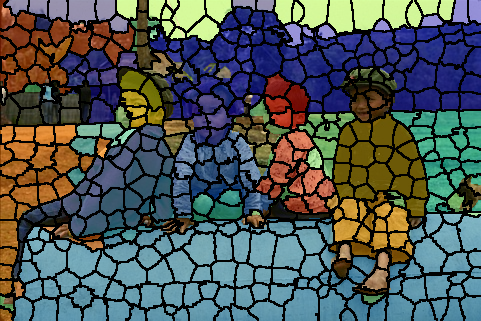}
\end{minipage}
\begin{minipage}[t]{0.135 \linewidth}
\includegraphics[scale=0.145]{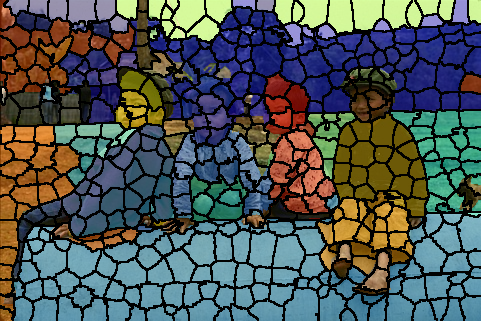}
\end{minipage} \\
\end{center}
\vspace{-11 pt}

\begin{center}
\begin{minipage}[t]{0.135 \linewidth}
\subfloat[]{\includegraphics[scale=0.145]{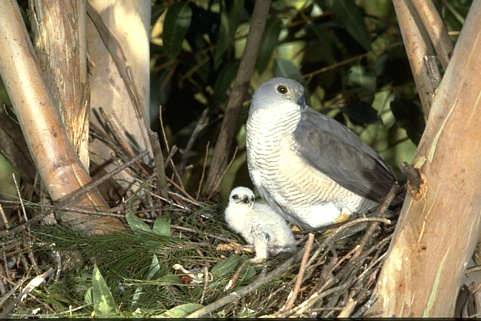}}
\end{minipage}
\begin{minipage}[t]{0.135 \linewidth}
\subfloat[]{\includegraphics[scale=0.145]{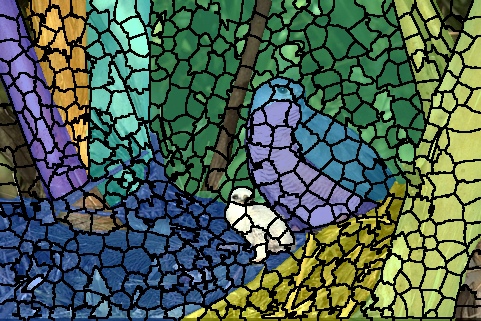}}
\end{minipage}
\begin{minipage}[t]{0.135 \linewidth}
\subfloat[]{\includegraphics[scale=0.145]{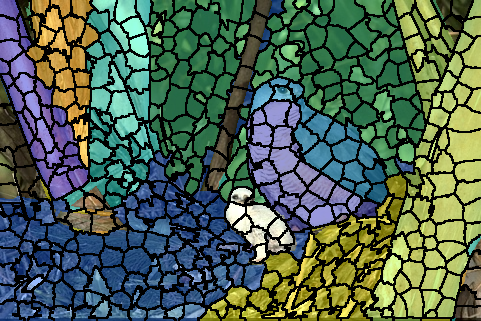}}
\end{minipage}
\begin{minipage}[t]{0.135 \linewidth}
\subfloat[]{\includegraphics[scale=0.145]{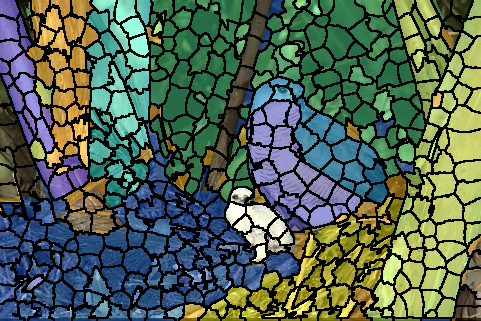}}
\end{minipage}
\begin{minipage}[t]{0.135 \linewidth}
\subfloat[]{\includegraphics[scale=0.145]{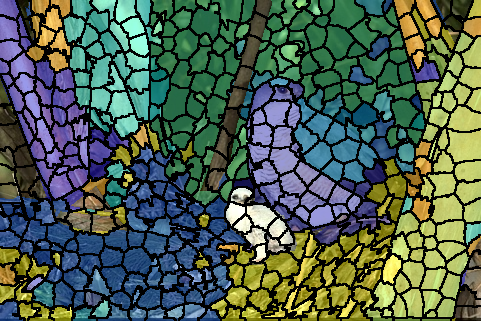}}
\end{minipage}
\begin{minipage}[t]{0.135 \linewidth}
\subfloat[]{\includegraphics[scale=0.145]{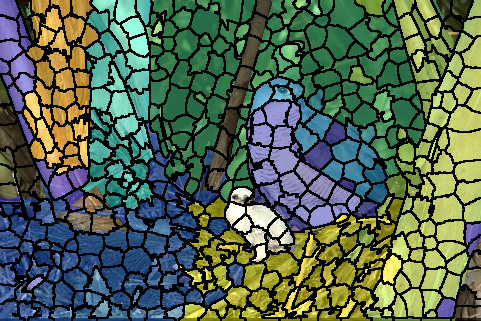}}
\end{minipage}
\begin{minipage}[t]{0.135 \linewidth}
\subfloat[]{\includegraphics[scale=0.145]{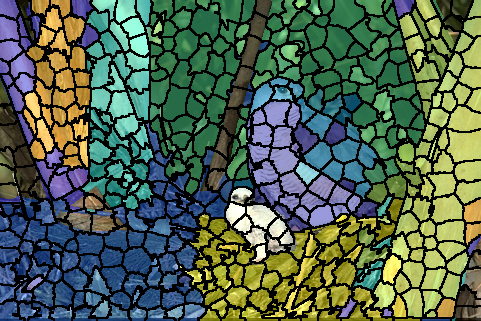}}
\end{minipage} \\
\end{center}
\vspace{-10 pt}

\caption{The performance of the prosed method on two relatively difficult images from BSDS 500 database. In all rows, a) the original image, b) ground truth, c) to g) are the outputs of 
DGL$_{GT}^{100 \%}$, DGL$_{GT}^{50 \%}$, DGL$_{GT}^{15 \hspace{1 pt} pts}$, DGL$_{BB}^{50 \%}$, DGL$_{BB}^{10 \% pt}$, respectively.}

\end{figure*}

\vspace{-10 pt}
\section{Simulations}
\vspace{-3 pt}
The proposed segmentation method is evaluated on the test images of Berkeley's BSDS 500 database \cite{Berkeley}. This set consists of 200 natural RGB 
images with multiple ground truth annotations. The images have sizes $321 \times 481$ where each color intensity is represented with 8 bits i.e.  
$|{\cal X}|=256^3$. As an accuracy measure we have considered intersection over union (IoU) which is the ratio of correctly labeled pixels to all labeled pixels (we considered 90\% of the labeled
pixels due to oversegmentations in images). We have estimated IoU by averaging it over multiple images and multiple ground truths.

Let GT$_i$ and BB$_i$, $i=1,2,...,M$, denote the ground truth masks and (tight) bounding boxes of 
$\Omega_i$. First, we have formed ${\cal T}_i$ from a percentage $f \%$, $f \in \{ 25, 50, 75, 100\}$, of randomly selected labeled pixels from GT$_i$ and BB$_i$.
We denote these methods by DGL$_{GT}^{f \%}$ and   DGL$_{BB}^{f\%}$, respectively. Next, we have considered the case
where $t$, $ t \in \{10 ,15, 20 \}$, random pixels seeds in $\Omega_i$ are selected and the case where bounding boxes are perturbed by $p \%$, $ p \in \{5 ,10, 15 \}$. These methods are denoted by DGL$_{GT}^{t \hspace{1 pt} pts}$ and  DGL$_{BB}^{p \% pt}$, respectively. In the former, ${\cal T}_i$ is constructed from the pixels within a square whose center is the seed point in $\Omega_i$ and whose side-length is $l$ (we used $l$=50 for BSDS images). In the latter approach, the two corner points, ($r_1,c_1$), ($r_2,c_2$), of bounding boxes that touch the diagonal are uniformly perturbed $p \%$ to obtain ($\hat{r}_1,\hat{c}_1$), ($\hat{r}_2,\hat{c}_2)$, respectively. Here, $\hat{r}_1$ is selected uniformly over $[ r1 - \frac{(r_2-r_1)p}{200}, r1 + \frac{(r_2-r_1)p}{200}]$ and we followed the same approached to obtain $\hat{c}_1, \hat{r}_2$ and $\hat{c}_2$.

The implementation of the proposed algorithm in RGB color space requires a memory proportional to $M^3 \times 256^3$ which becomes unpractical for images with $M>10$. Thus, we used 
dimensionality reduction by implementing the proposed algorithm in the HSV color space and considering only H and S components of the image. This approach has the advantage of decoupling the chromatic information from the shading effect \cite{Mignotte, Gonzalez} and the resultant algorithm showed negligible performance degradation when $|{\cal X}_1|=|{\cal X}_2|=1024$  where  ${\cal X}_1$ and ${\cal X}_2$ correspond to equidistant bins for H and S components, respectively. The performance of this approach is demonstrated in Figure 3 for two relatively difficult images where we have
used the SLIC framework \cite{SLIC} with $K=500$ superpixels. Inspecting the images we observe that mislabeling occurs when i)  regions with similar intensity distributions are annotated differently due to the presence of a contour e.g. the sky in the top image and tree branches in the bottom image ii) the intensity distributions mix while construction ${\cal T}_i$ in  DGL$_{GT}^{t \hspace{1 pt} pts}$, DGL$_{BB}^{f\%}$ and DGL$_{BB}^{p \% pt}$ e.g. the mislabeled superpixels in the body and in the wing of the mother bird.

For benchmarking we have reduced the alphabet size and used the algorithm with $|{\cal X}_1|=|{\cal X}_2|=\sqrt{321 \times 482} \approx 392$ so that it has linear complexity in the number of pixels. This reduction resulted in some degradation, around 3\%-5\%,  in the accuracy. We have also investigated the effect of additional user inputs by assuming a genie-aided user optimally relabels the mislabeled (or partially mislabeled) superpixels which can be performed by at most two point clicks per superpixel. With this approach, the amount of increase in the accuracy per additional user inputs (clicks) can be easily calculated during benchmarking without actually performing the corrections.  

The simulation results are presented in Figure 4. Recall that ${\cal T}_i$ consists of labeled
pixels from $\Omega_i$ in DGL$_{GT}^{f \%}$, therefore DGL$_{GT}^{100 \%}$ exhibits the ultimate performance of the proposed method. Inspecting Figure 4, we observe that the performances of DGL$_{GT}^{100 \%}$, DGL$_{GT}^{75 \%}$ and DGL$_{GT}^{50\%}$ are almost the same and one can reach an initial segmentation accuracy around 0.9 when only half of the labeled pixels in ground truths masks are provided by the user. This performance is promising given the simplicity of the proposed method. As expected from the discussion in Section 2C, the performance deteriorates in DGL$_{BB}^{f\%}$, DGL$_{GT}^{t \hspace{1 pt} pts}$ and DGL$_{BB}^{p \% pt}$. However, this deterioration is consistent in each of the considered user input type which indicates the robustness of the proposed method.
  
\begin{figure}[t]
\begin{center}
 \includegraphics[scale=0.22]{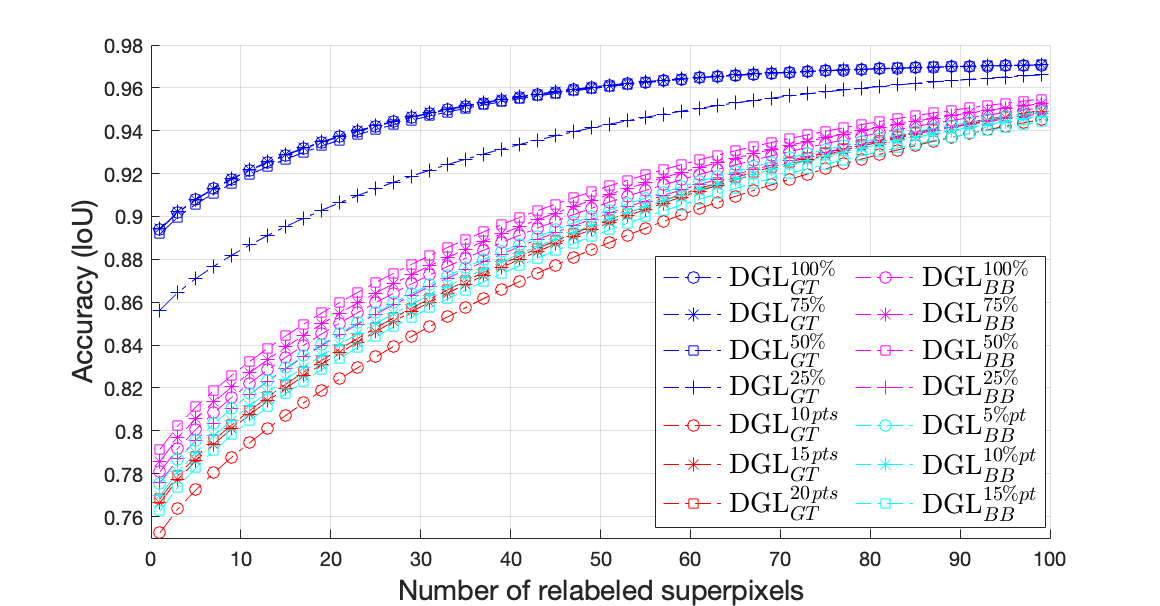}
  \end{center}
   \vspace{-8 pt} 
  \caption{Benchmarking results on BSDS 500 database}
   \end{figure}
  
 \vspace{-10 pt} 
 \section{Conclusion}
  \vspace{-5 pt} 
  
We have presented a novel, user-assisted, multiple segmentation method based on DGL testing procedure. The proposed method is shown to be robust when used with
different user input types such as labeled pixels and seeds from ground truth masks or bounding boxes that may be perturbed. Having linear time complexity in the number of pixels and requiring
minimal coding, around 30 lines, the proposed method is suitable for fast benchmarking purposes. In this work, we have only utilized the empirical distributions, i.e. histograms, in the user input regions.
Including the spatial information of these regions, e.g.  as in \cite{Cremers}, in the DGL test may improve the performance of the proposed method and is the topic of our future work.

\end{document}

%% file: fig_0.tex

\tikzset{every picture/.style={line width=0.75pt}} 

\begin{tikzpicture}[x=0.75pt,y=0.75pt,yscale=-0.60,xscale=0.60]
\draw  [color=black ,draw opacity=1 ][line width=1.5]  (27,20) -- (258,20) -- (258,217) -- (27,217) -- cycle ;
\draw  [color=black ,draw opacity=1 ][line width=1.5]  (48,78) .. controls (48,52.59) and (63.67,32) .. (83,32) .. controls (102.33,32) and (118,52.59) .. (118,78) .. controls (118,103.41) and (102.33,124) .. (83,124) .. controls (63.67,124) and (48,103.41) .. (48,78) -- cycle ;
\draw  [color=black ,draw opacity=1 ][line width=1.5]  (200,28) -- (235,63) -- (200,98) -- (165,63) -- cycle ;
\draw  [color=black ,draw opacity=1 ] [line width=1.5]  (179,119) -- (254,119) -- (254,202) -- (179,202) -- cycle ;
\draw  [color=black ,draw opacity=1 ][line width=1.5]  (39,171.5) .. controls (39,156.31) and (66.98,144) .. (101.5,144) .. controls (136.02,144) and (164,156.31) .. (164,171.5) .. controls (164,186.69) and (136.02,199) .. (101.5,199) .. controls (66.98,199) and (39,186.69) .. (39,171.5) -- cycle ;

\draw (72,159) node [anchor=north west][inner sep=0.75pt, scale=0.8]   [align=left]  {$\Omega_2; P_2$};
\draw (59,67) node [anchor=north west][inner sep=0.75pt, scale=0.8]   [align=left] {$\Omega_1; P_1$};
\draw (175,53) node [anchor=north west][inner sep=0.75pt, scale=0.8]   [align=left]{$\Omega_2; P_2$};
\draw (187,150) node [anchor=north west][inner sep=0.75pt, scale=0.8]   [align=left] {$\Omega_3; P_3$};
\draw (121,96) node [anchor=north west][inner sep=0.75pt, scale=0.8]   [align=left]  {$\Omega_4; P_4$};

\end{tikzpicture}

%% file: fig_1.tex
\tikzset{every picture/.style={line width=0.75pt}} 

\begin{tikzpicture}[x=0.75pt,y=0.75pt,yscale=-0.6,xscale=0.6]

\draw  [dash pattern={on 4.5pt off 4.5pt}] [line width=0.50] (27,20) -- (47,20) -- (55,53) -- (26,53) -- cycle ;
\draw  [dash pattern={on 4.5pt off 4.5pt}] [line width=0.50]  (47,20) -- (76,20) -- (74,32) -- (55,53) -- cycle ;
\draw  [dash pattern={on 4.5pt off 4.5pt}] [line width=0.50]  (76,20) -- (105,20) -- (97,35) -- (74,32) -- cycle ;
\draw [dash pattern={on 4.5pt off 4.5pt}] [line width=0.50]  (105,20) -- (134,20) -- (133,42) -- (97,35) -- cycle ;
\draw [dash pattern={on 4.5pt off 4.5pt}]  [line width=0.50]  (134,20) -- (170,20) -- (174,42) -- (133,42) -- cycle ;
\draw  [dash pattern={on 4.5pt off 4.5pt}] [line width=0.50]  (26,53) -- (55,53) -- (47,87) -- (26,87) -- cycle ;
\draw  [dash pattern={on 4.5pt off 4.5pt}] [line width=0.50]  (26,87) -- (47,87) -- (65,119) -- (26,121) -- cycle ;
\draw  [dash pattern={on 4.5pt off 4.5pt}]  [line width=0.50] (26,121) -- (65,119) -- (58,152) -- (26,155) -- cycle ;
\draw [dash pattern={on 4.5pt off 4.5pt}]  [line width=0.50] (26,155) -- (58,152) -- (42,180) -- (27,184) -- cycle ;
\draw  [dash pattern={on 4.5pt off 4.5pt}]  [line width=0.50] (27,184) -- (42,180) -- (46,218) -- (27,217) -- cycle ;
\draw  [dash pattern={on 4.5pt off 4.5pt}] [line width=0.50]  (42,180) -- (72,195) -- (78,217) -- (46,218) -- cycle ;
\draw [dash pattern={on 4.5pt off 4.5pt}]  [line width=0.50] (65,119) -- (104,117) -- (101.5,144) -- (58,152) -- cycle ;
\draw  [dash pattern={on 4.5pt off 4.5pt}]  [line width=0.50] (104,117) -- (119,90) -- (140,150) -- (101.5,144) -- cycle ;
\draw [dash pattern={on 4.5pt off 4.5pt}] [line width=0.50]  (118,69) -- (136,69) -- (139,96) -- (115.5,96) -- cycle ;
\draw [dash pattern={on 4.5pt off 4.5pt}] [line width=0.50]  (170,20) -- (193,20) -- (200,28) -- (177,53) -- cycle ;
\draw [dash pattern={on 4.5pt off 4.5pt}]  [line width=0.50] (97,35) -- (133,42) -- (136,69) -- (118,69) -- cycle ;
\draw [dash pattern={on 4.5pt off 4.5pt}] [line width=0.50]  (218,21) -- (258,20) -- (260,54) -- (237,50) -- cycle ;
\draw  [dash pattern={on 4.5pt off 4.5pt}]  [line width=0.50] (133,42) -- (174,42) -- (177,53) -- (136,69) -- cycle ;
\draw [dash pattern={on 4.5pt off 4.5pt}] [line width=0.50]  (136,69) -- (177,53) -- (171,68) -- (139,96) -- cycle ;
\draw [dash pattern={on 4.5pt off 4.5pt}]  [line width=0.50] (193,20) -- (218,21) -- (237,50) -- (222,49) -- cycle ;
\draw  [dash pattern={on 4.5pt off 4.5pt}]  [line width=0.50] (121,96) -- (139,96) -- (159,116) -- (140,150) -- cycle ;
\draw  [dash pattern={on 4.5pt off 4.5pt}] [line width=0.50]  (222,49) -- (260,54) -- (258,99) -- (235,63) -- cycle ;
\draw  [dash pattern={on 4.5pt off 4.5pt}] [line width=0.50]  (200,98) -- (235,63) -- (258,99) -- (232,105) -- cycle ;
\draw  [dash pattern={on 4.5pt off 4.5pt}]  [line width=0.50] (140,150) -- (159,116) -- (188.5,119.25) -- (174,163) -- cycle ;
\draw  [dash pattern={on 4.5pt off 4.5pt}]  [line width=0.50] (139,96) -- (171,68) -- (199.5,99) -- (159,116) -- cycle ;
\draw  [dash pattern={on 4.5pt off 4.5pt}]  [line width=0.50] (163,115.5) -- (200,98) -- (232,105) -- (214,123) -- cycle ;
\draw  [dash pattern={on 4.5pt off 4.5pt}]  [line width=0.50] (232,105) -- (258,99) -- (260,122) -- (214,123) -- cycle ;
\draw  [dash pattern={on 4.5pt off 4.5pt}]  [line width=0.50] (72,195) -- (112,197) -- (118,217) -- (79,218) -- cycle ;
\draw  [dash pattern={on 4.5pt off 4.5pt}]  [line width=0.50] (110.5,196.5) -- (152,187) -- (156,217) -- (120,217) -- cycle ;
\draw  [dash pattern={on 4.5pt off 4.5pt}] [line width=0.50]  (152,187) -- (174,163) -- (183,217) -- (156,217) -- cycle ;
\draw  [dash pattern={on 4.5pt off 4.5pt}]  [line width=0.50] (182,197) -- (254,202) -- (258,217) -- (183,218) -- cycle ;
\draw  [dash pattern={on 4.5pt off 4.5pt}]  [line width=0.50] (53,155) -- (79.75,148) -- (72,195) -- (42,180) -- cycle ;
\draw  [dash pattern={on 4.5pt off 4.5pt}] [line width=0.50]  (75.88,171.5) -- (115,180) -- (112,197) -- (72,195) -- cycle ;
\draw  [dash pattern={on 4.5pt off 4.5pt}]  [line width=0.50] (115,180) -- (145,171) -- (154.5,188.5) -- (112,197) -- cycle ;
\draw [dash pattern={on 4.5pt off 4.5pt}]  [line width=0.50] (145,171) -- (162,159) -- (174,163) -- (152,187) -- cycle ;
\draw [dash pattern={on 4.5pt off 4.5pt}]  [line width=0.50] (79.75,148) -- (101.5,144) -- (115,180) -- (75.88,171.5) -- cycle ;
\draw [dash pattern={on 4.5pt off 4.5pt}]  [line width=0.50] (101.5,144) -- (140,150) -- (145,171) -- (115,180) -- cycle ;
\draw  [dash pattern={on 4.5pt off 4.5pt}]  [line width=0.50] (140,150) -- (162,159) -- (145,171) -- cycle ;
\draw  [dash pattern={on 4.5pt off 4.5pt}]  [line width=0.50] (177,53) -- (217.5,80.5) -- (199.5,99) -- (169.75,66.5) -- cycle ;
\draw  [dash pattern={on 4.5pt off 4.5pt}]  [line width=0.50] (200,28) -- (235,63) -- (212.13,83) -- (177,53) -- cycle ;
\draw  [dash pattern={on 4.5pt off 4.5pt}]  [line width=0.50] (74,32) -- (83,78) -- (47,87) -- (51,57) -- cycle ;
\draw  [dash pattern={on 4.5pt off 4.5pt}]  [line width=0.50] (83,78) -- (90,101) -- (65,119) -- (47,87) -- cycle ;
\draw [dash pattern={on 4.5pt off 4.5pt}]  [line width=0.50] (90,101) -- (111.5,103.5) -- (104,117) -- (65,119) -- cycle ;
\draw  [dash pattern={on 4.5pt off 4.5pt}]  [line width=0.50] (83,78) -- (118,78) -- (111.5,103.5) -- (90,101) -- cycle ;
\draw  [dash pattern={on 4.5pt off 4.5pt}]  [line width=0.50] (74,32) -- (106,38) -- (118,78) -- (83,78) -- cycle ;
\draw  [color=black ,draw opacity=1 ][line width=1.5]  (48,78) .. controls (48,52.59) and (63.67,32) .. (83,32) .. controls (102.33,32) and (118,52.59) .. (118,78) .. controls (118,103.41) and (102.33,124) .. (83,124) .. controls (63.67,124) and (48,103.41) .. (48,78) -- cycle ;
\draw  [color=black  ,draw opacity=1 ][line width=1.5]  (39,171.5) .. controls (39,156.31) and (66.98,144) .. (101.5,144) .. controls (136.02,144) and (164,156.31) .. (164,171.5) .. controls (164,186.69) and (136.02,199) .. (101.5,199) .. controls (66.98,199) and (39,186.69) .. (39,171.5) -- cycle ;
\draw  [color=black ,draw opacity=1 ][line width=1.5]  (200,28) -- (235,63) -- (200,98) -- (165,63) -- cycle ;
\draw  [dash pattern={on 4.5pt off 4.5pt}] (174,163) -- (207,162) -- (218,199.5) -- (182,197) -- cycle ;
\draw  [dash pattern={on 4.5pt off 4.5pt}] (207,162) -- (254.5,164.75) -- (254,202) -- (218,199.5) -- cycle ;
\draw [dash pattern={on 4.5pt off 4.5pt}] (208,122) -- (254,119) -- (254.5,164.75) -- (212.75,162.13) -- cycle ;
\draw   [dash pattern={on 4.5pt off 4.5pt}] (188.5,119.25) -- (208,122) -- (212.75,162.13) -- (174,163) -- cycle ;
\draw   [color=black  ,draw opacity=1 ] [line width=1.5]  (179,119) -- (254,119) -- (254,202) -- (179,202) -- cycle ;
\draw  [color={rgb, 255:red, 0; green, 0; blue, 0 }  ,draw opacity=1 ][line width=1.5]  (27,20) -- (258,20) -- (258,217) -- (27,217) -- cycle ;

\draw (82,159) node [text=black, anchor=north west][inner sep=0.75pt, scale=1.2]   [align=left] {$\Omega_2$};
\draw (69,67) node [text=black,anchor=north west][inner sep=0.75pt, scale=1.2]   [align=left] {$\Omega_1$};
\draw (180,53) node [text=black, anchor=north west][inner sep=0.75pt, scale=1.2]   [align=left] {$\Omega_2$};
\draw (197,150) node [text=black, anchor=north west][inner sep=0.75pt, scale=1.2]   [align=left] {$\Omega_3$};
\draw (131,96) node [text=black, anchor=north west][inner sep=0.75pt, scale=1.2]   [align=left] {$\Omega_4$};

\end{tikzpicture}

%% file: fig_2.tex
\tikzset{every picture/.style={line width=0.75pt}} 

\begin{tikzpicture}[x=0.75pt,y=0.75pt,yscale=-0.6,xscale=0.6]

\draw  [fill={rgb, 255:red, 248; green, 231; blue, 28 }  ,fill opacity=1 ][dash pattern={on 4.5pt off 4.5pt}] (27,20) -- (47,20) -- (55,53) -- (26,53) -- cycle ;
\draw  [fill={rgb, 255:red, 248; green, 231; blue, 28 }  ,fill opacity=1 ][dash pattern={on 4.5pt off 4.5pt}] (47,20) -- (76,20) -- (74,32) -- (55,53) -- cycle ;
\draw  [fill={rgb, 255:red, 248; green, 231; blue, 28 }  ,fill opacity=1 ][dash pattern={on 4.5pt off 4.5pt}] (76,20) -- (105,20) -- (97,35) -- (74,32) -- cycle ;
\draw  [fill={rgb, 255:red, 248; green, 231; blue, 28 }  ,fill opacity=1 ][dash pattern={on 4.5pt off 4.5pt}] (105,20) -- (134,20) -- (133,42) -- (97,35) -- cycle ;
\draw  [fill={rgb, 255:red, 248; green, 231; blue, 28 }  ,fill opacity=1 ][dash pattern={on 4.5pt off 4.5pt}] (134,20) -- (170,20) -- (174,42) -- (133,42) -- cycle ;
\draw  [fill={rgb, 255:red, 248; green, 231; blue, 28 }  ,fill opacity=1 ][dash pattern={on 4.5pt off 4.5pt}] (26,53) -- (55,53) -- (47,87) -- (26,87) -- cycle ;
\draw  [fill={rgb, 255:red, 248; green, 231; blue, 28 }  ,fill opacity=1 ][dash pattern={on 4.5pt off 4.5pt}] (26,87) -- (47,87) -- (65,119) -- (26,121) -- cycle ;
\draw  [fill={rgb, 255:red, 248; green, 231; blue, 28 }  ,fill opacity=1 ][dash pattern={on 4.5pt off 4.5pt}] (26,121) -- (65,119) -- (58,152) -- (26,155) -- cycle ;
\draw  [fill={rgb, 255:red, 248; green, 231; blue, 28 }  ,fill opacity=1 ][dash pattern={on 4.5pt off 4.5pt}] (26,155) -- (58,152) -- (42,180) -- (27,184) -- cycle ;
\draw  [fill={rgb, 255:red, 248; green, 231; blue, 28 }  ,fill opacity=1 ][dash pattern={on 4.5pt off 4.5pt}] (27,184) -- (42,180) -- (46,218) -- (27,217) -- cycle ;
\draw  [fill={rgb, 255:red, 248; green, 231; blue, 28 }  ,fill opacity=1 ][dash pattern={on 4.5pt off 4.5pt}] (42,180) -- (72,195) -- (78,217) -- (46,218) -- cycle ;
\draw  [fill={rgb, 255:red, 248; green, 231; blue, 28 }  ,fill opacity=1 ][dash pattern={on 4.5pt off 4.5pt}] (65,119) -- (104,117) -- (101.5,144) -- (58,152) -- cycle ;
\draw  [fill={rgb, 255:red, 248; green, 231; blue, 28 }  ,fill opacity=1 ][dash pattern={on 4.5pt off 4.5pt}] (104,117) -- (119,90) -- (140,150) -- (101.5,144) -- cycle ;
\draw  [fill={rgb, 255:red, 248; green, 231; blue, 28 }  ,fill opacity=1 ][dash pattern={on 4.5pt off 4.5pt}] (118,69) -- (136,69) -- (139,96) -- (115.5,96) -- cycle ;
\draw  [fill={rgb, 255:red, 248; green, 231; blue, 28 }  ,fill opacity=1 ][dash pattern={on 4.5pt off 4.5pt}] (170,20) -- (193,20) -- (200,28) -- (177,53) -- cycle ;
\draw  [fill={rgb, 255:red, 248; green, 231; blue, 28 }  ,fill opacity=1 ][dash pattern={on 4.5pt off 4.5pt}] (97,35) -- (133,42) -- (136,69) -- (118,69) -- cycle ;
\draw  [fill={rgb, 255:red, 248; green, 231; blue, 28 }  ,fill opacity=1 ][dash pattern={on 4.5pt off 4.5pt}] (218,21) -- (258,20) -- (260,54) -- (237,50) -- cycle ;
\draw  [fill={rgb, 255:red, 248; green, 231; blue, 28 }  ,fill opacity=1 ][dash pattern={on 4.5pt off 4.5pt}] (133,42) -- (174,42) -- (177,53) -- (136,69) -- cycle ;
\draw  [fill={rgb, 255:red, 248; green, 231; blue, 28 }  ,fill opacity=1 ][dash pattern={on 4.5pt off 4.5pt}] (136,69) -- (177,53) -- (171,68) -- (139,96) -- cycle ;
\draw  [fill={rgb, 255:red, 248; green, 231; blue, 28 }  ,fill opacity=1 ][dash pattern={on 4.5pt off 4.5pt}] (193,20) -- (218,21) -- (237,50) -- (222,49) -- cycle ;
\draw  [fill={rgb, 255:red, 248; green, 231; blue, 28 }  ,fill opacity=1 ][dash pattern={on 4.5pt off 4.5pt}] (121,96) -- (139,96) -- (159,116) -- (140,150) -- cycle ;
\draw  [fill={rgb, 255:red, 248; green, 231; blue, 28 }  ,fill opacity=1 ][dash pattern={on 4.5pt off 4.5pt}] (222,49) -- (260,54) -- (258,99) -- (235,63) -- cycle ;
\draw  [fill={rgb, 255:red, 248; green, 231; blue, 28 }  ,fill opacity=1 ][dash pattern={on 4.5pt off 4.5pt}] (200,98) -- (235,63) -- (258,99) -- (232,105) -- cycle ;
\draw  [fill={rgb, 255:red, 248; green, 231; blue, 28 }  ,fill opacity=1 ][dash pattern={on 4.5pt off 4.5pt}] (140,150) -- (159,116) -- (188.5,119.25) -- (174,163) -- cycle ;
\draw  [fill={rgb, 255:red, 248; green, 231; blue, 28 }  ,fill opacity=1 ][dash pattern={on 4.5pt off 4.5pt}] (139,96) -- (171,68) -- (199.5,99) -- (159,116) -- cycle ;
\draw  [fill={rgb, 255:red, 248; green, 231; blue, 28 }  ,fill opacity=1 ][dash pattern={on 4.5pt off 4.5pt}] (163,115.5) -- (200,98) -- (232,105) -- (214,123) -- cycle ;
\draw  [fill={rgb, 255:red, 248; green, 231; blue, 28 }  ,fill opacity=1 ][dash pattern={on 4.5pt off 4.5pt}] (232,105) -- (258,99) -- (260,122) -- (214,123) -- cycle ;
\draw  [fill={rgb, 255:red, 248; green, 231; blue, 28 }  ,fill opacity=1 ][dash pattern={on 4.5pt off 4.5pt}] (72,195) -- (112,197) -- (118,217) -- (79,218) -- cycle ;
\draw  [fill={rgb, 255:red, 248; green, 231; blue, 28 }  ,fill opacity=1 ][dash pattern={on 4.5pt off 4.5pt}] (110.5,196.5) -- (152,187) -- (156,217) -- (120,217) -- cycle ;
\draw  [fill={rgb, 255:red, 248; green, 231; blue, 28 }  ,fill opacity=1 ][dash pattern={on 4.5pt off 4.5pt}] (152,187) -- (174,163) -- (183,217) -- (156,217) -- cycle ;
\draw  [fill={rgb, 255:red, 248; green, 231; blue, 28 }  ,fill opacity=1 ][dash pattern={on 4.5pt off 4.5pt}] (182,197) -- (254,202) -- (258,217) -- (183,218) -- cycle ;
\draw  [fill={rgb, 255:red, 245; green, 166; blue, 35 }  ,fill opacity=1 ][dash pattern={on 4.5pt off 4.5pt}] (53,155) -- (79.75,148) -- (72,195) -- (42,180) -- cycle ;
\draw  [fill={rgb, 255:red, 245; green, 166; blue, 35 }  ,fill opacity=1 ][dash pattern={on 4.5pt off 4.5pt}] (75.88,171.5) -- (115,180) -- (112,197) -- (72,195) -- cycle ;
\draw  [fill={rgb, 255:red, 245; green, 166; blue, 35 }  ,fill opacity=1 ][dash pattern={on 4.5pt off 4.5pt}] (115,180) -- (145,171) -- (154.5,188.5) -- (112,197) -- cycle ;
\draw  [fill={rgb, 255:red, 245; green, 166; blue, 35 }  ,fill opacity=1 ][dash pattern={on 4.5pt off 4.5pt}] (145,171) -- (162,159) -- (174,163) -- (152,187) -- cycle ;
\draw  [fill={rgb, 255:red, 245; green, 166; blue, 35 }  ,fill opacity=1 ][dash pattern={on 4.5pt off 4.5pt}] (79.75,148) -- (101.5,144) -- (115,180) -- (75.88,171.5) -- cycle ;
\draw  [fill={rgb, 255:red, 245; green, 166; blue, 35 }  ,fill opacity=1 ][dash pattern={on 4.5pt off 4.5pt}] (101.5,144) -- (140,150) -- (145,171) -- (115,180) -- cycle ;
\draw  [fill={rgb, 255:red, 245; green, 166; blue, 35 }  ,fill opacity=1 ][dash pattern={on 4.5pt off 4.5pt}] (140,150) -- (162,159) -- (145,171) -- cycle ;
\draw  [fill={rgb, 255:red, 245; green, 166; blue, 35 }  ,fill opacity=1 ][dash pattern={on 4.5pt off 4.5pt}] (177,53) -- (217.5,80.5) -- (199.5,99) -- (169.75,66.5) -- cycle ;
\draw  [fill={rgb, 255:red, 245; green, 166; blue, 35 }  ,fill opacity=1 ][dash pattern={on 4.5pt off 4.5pt}] (200,28) -- (235,63) -- (212.13,83) -- (177,53) -- cycle ;
\draw  [fill={rgb, 255:red, 126; green, 211; blue, 33 }  ,fill opacity=1 ][dash pattern={on 4.5pt off 4.5pt}] (74,32) -- (83,78) -- (47,87) -- (51,57) -- cycle ;
\draw  [fill={rgb, 255:red, 126; green, 211; blue, 33 }  ,fill opacity=1 ][dash pattern={on 4.5pt off 4.5pt}] (83,78) -- (90,101) -- (65,119) -- (47,87) -- cycle ;
\draw  [fill={rgb, 255:red, 126; green, 211; blue, 33 }  ,fill opacity=1 ][dash pattern={on 4.5pt off 4.5pt}] (90,101) -- (111.5,103.5) -- (104,117) -- (65,119) -- cycle ;
\draw  [fill={rgb, 255:red, 126; green, 211; blue, 33 }  ,fill opacity=1 ][dash pattern={on 4.5pt off 4.5pt}] (83,78) -- (118,78) -- (111.5,103.5) -- (90,101) -- cycle ;
\draw  [fill={rgb, 255:red, 126; green, 211; blue, 33 }  ,fill opacity=1 ][dash pattern={on 4.5pt off 4.5pt}] (74,32) -- (106,38) -- (118,78) -- (83,78) -- cycle ;
\draw  [color={rgb, 255:red, 0; green, 0; blue, 0 }  ,draw opacity=1 ][line width=1.5]  (48,78) .. controls (48,52.59) and (63.67,32) .. (83,32) .. controls (102.33,32) and (118,52.59) .. (118,78) .. controls (118,103.41) and (102.33,124) .. (83,124) .. controls (63.67,124) and (48,103.41) .. (48,78) -- cycle ;
\draw  [color={rgb, 255:red, 0; green, 0; blue, 0 }  ,draw opacity=1 ][line width=1.5]  (39,171.5) .. controls (39,156.31) and (66.98,144) .. (101.5,144) .. controls (136.02,144) and (164,156.31) .. (164,171.5) .. controls (164,186.69) and (136.02,199) .. (101.5,199) .. controls (66.98,199) and (39,186.69) .. (39,171.5) -- cycle ;
\draw  [color={rgb, 255:red, 0; green, 0; blue, 0 }  ,draw opacity=1 ][line width=1.5]  (200,28) -- (235,63) -- (200,98) -- (165,63) -- cycle ;
\draw  [fill={rgb, 255:red, 80; green, 227; blue, 194 }  ,fill opacity=1 ][dash pattern={on 4.5pt off 4.5pt}] (174,163) -- (207,162) -- (218,199.5) -- (182,197) -- cycle ;
\draw  [fill={rgb, 255:red, 80; green, 227; blue, 194 }  ,fill opacity=1 ][dash pattern={on 4.5pt off 4.5pt}] (207,162) -- (254.5,164.75) -- (254,202) -- (218,199.5) -- cycle ;
\draw  [fill={rgb, 255:red, 80; green, 227; blue, 194 }  ,fill opacity=1 ][dash pattern={on 4.5pt off 4.5pt}] (208,122) -- (254,119) -- (254.5,164.75) -- (212.75,162.13) -- cycle ;
\draw  [fill={rgb, 255:red, 80; green, 227; blue, 194 }  ,fill opacity=1 ][dash pattern={on 4.5pt off 4.5pt}] (188.5,119.25) -- (208,122) -- (212.75,162.13) -- (174,163) -- cycle ;
\draw  [line width=1.5]  (179,119) -- (254,119) -- (254,202) -- (179,202) -- cycle ;
\draw  [color={rgb, 255:red, 0; green, 0; blue, 0 }  ,draw opacity=1 ][line width=1.5]  (27,20) -- (258,20) -- (258,217) -- (27,217) -- cycle ;

\draw (82,159) node [anchor=north west][inner sep=0.75pt, scale=1.2]   [align=left] {$\Omega_2$};
\draw (69,67) node [anchor=north west][inner sep=0.75pt, scale=1.2]   [align=left] {$\Omega_1$};
\draw (180,53) node [anchor=north west][inner sep=0.75pt, scale=1.2]   [align=left] {$\Omega_2$};
\draw (197,150) node [anchor=north west][inner sep=0.75pt, scale=1.2]   [align=left] {$\Omega_3$};
\draw (131,96) node [anchor=north west][inner sep=0.75pt, scale=1.2]   [align=left] {$\Omega_4$};

\end{tikzpicture}